# Discovery of partial differential equations from highly noisy and sparse data with physics-informed information criterion

Hao Xu[a], Junsheng Zeng[b], and Dongxiao Zhang[c,d,*]

[a] *BIC-ESAT, ERE, and SKLTCS, College of Engineering, Peking University, Beijing 100871, P. R. China*

[b] *Institute of Applied Physics and Computational Mathematics, Beijing 100088, P. R. China*

[c] *National Center for Applied Mathematics Shenzhen (NCAMS), Southern University of Science and Technology, Shenzhen 518055, Guangdong, P. R. China*

[d] *Department of Mathematics and Theories, Peng Cheng Laboratory, Shenzhen 518000, Guangdong, P. R. China*

[*] Corresponding author.

E-mail address: 390260267@pku.edu.cn (H. Xu); zengjs1993@163.com (J. Zeng); zhangdx@sustech.edu.cn (D. Zhang).

**Abstract:** Data-driven discovery of PDEs has made tremendous progress recently, and many canonical PDEs have been discovered successfully for proof-of-concept. However, determining the most proper PDE without prior references remains challenging in terms of practical applications. In this work, a physics-informed information criterion (PIC) is proposed to measure the parsimony and precision of the discovered PDE synthetically. The proposed PIC achieves state-of-the-art robustness to highly noisy and sparse data on seven canonical PDEs from different physical scenes, which confirms its ability to handle difficult situations. The PIC is also employed to discover unrevealed macroscale governing equations from microscopic simulation data in an actual physical scene. The results show that the discovered macroscale PDE is precise and parsimonious, and satisfies underlying symmetries, which facilitates understanding and simulation of the physical process. The proposition of PIC enables practical applications of PDE discovery in discovering unrevealed governing equations in broader physical scenes.

**Keywords**: PDE discovery; information criterion; physics-informed neural network; physics-informed information criterion.

With the development of machine learning techniques and computational power, the data-driven discovery of partial differential equations (PDEs) has made tremendous progress over the past years. Unlike the paradigm of deriving physical laws from first principles or constitution laws, it provides an alternative path by directly discovering governing equations from observation data, which is more appropriate and practical for systems with elusive underlying mechanisms.

In essence, the discovery of PDEs constitutes a sparse regression task in which a few terms with non-zero coefficients are identified from unlimited candidates. In the early stage of research, sparse regression methods, including Lasso[1], sequential threshold ridge regression (STRidge)[2] and SINDy[3], were adopted to discover PDEs from a limited candidate library given beforehand. The following works have introduced various techniques to improve the above-mentioned methods to deal with high noise and complex systems[4–9]. Sparse regression-based methods have the advantage of high calculation efficiency, but applications are limited by the requirement of the pre-determined complete candidate library. Therefore, the genetic algorithm is employed to discover PDEs in the incomplete library, since unlimited combinations can be generated from several basic genes by cross-over and mutation[10,11]. The utilization of symbolic regression further relaxes the restrictions of the candidate library and makes the discovery of free-form PDEs possible[12]. The rapid development of deep learning techniques has brought additional possibilities to PDE discovery. Automatic differentiation of the neural network has significantly improved the flexibility and accuracy of derivatives calculation, which assists in solving the problem of sparse data and high noise[13–15]. Among neural network-based techniques, the physics-informed neural network (PINN)[16] shows excellent robustness and accuracy when identifying PDEs from data that are sparse and with high levels of noise[17–19]. However, the computational cost of PINN is high, and the results' accuracy is highly dependent on physical constraints.

Most of the methods mentioned above are proposed for proof-of-concept, where the data are originated from the numerical solution of canonical PDEs or systems. Although discovering canonical PDEs is an excellent way to examine the ability of the proposed methods, problems will emerge regarding practical applications. The core issue is how to determine whether the discovered PDE is correct (or proper). In the stage of proof-of-concept, this is not a significant problem since the discovered PDE can be compared with the true PDE that is already known. However, in practical applications, the target is to find an unrevealed PDE that can describe the physical process well, while no referenced PDE is provided. To solve this problem, Zhang and Liu[15] attempted to solve the discovered PDE numerically and compare it with observation data to determine the optimal PDE. However, solving PDEs is time-consuming, and some potentially discovered PDEs are difficult to solve using numerical methods without extensive research because of nonlinearity and instability. Meanwhile, the comparison with observation data can only measure the precision of the discovered PDE, but parsimony is neglected. For example, a PDE with redundant compensation terms may have a smaller numerical error, but it is not the simplest form, which makes it hard to explain the physical meaning. Furthermore, the outcomes of existing methods are often influenced by the selection of core hyper-parameters, including the magnitude of regularization and thresholds, which is unfavorable in practical applications in which the hyper-parameters are difficult to adjust without a referenced PDE.

In general, a proper PDE for describing a physical process should satisfy three principles: precision; parsimony; and interpretability. Inspired by these principles, a brand-new information criterion, called the physics-informed information criterion (PIC), is proposed in this work, which combines the measurement of precision and parsimony to select a proper PDE with physical interpretability. In the PIC, the moving horizon technique is introduced to identify redundant terms and measure the parsimony. At the same time, the PINN method is employed to estimate the precision without solving potential PDEs numerically.

To demonstrate the robustness and accuracy of PIC, we apply it to identify seven canonical

PDEs from different physical fields under high levels of noise. We also compare the proposed PIC with commonly-used information criteria, such as the Akaike information criterion (AIC) and the Bayesian information criterion (BIC), to demonstrate the superiority of PIC faced with high noise. The PIC is also compared with sparse regression-based methods, including Lasso and STRidge, which utilize $L_1$ and $L_2$ regularization with the threshold to maintain parsimony, respectively. Finally, and importantly, we adopt the PIC criterion to discover unrevealed PDEs from an actual physical scene, in which the macroscale governing equations have not yet been explicitly proposed.

**Results**

**Overview of the PIC criterion.** In this work, the mathematical form of the PDE discovery problem is written as follows:

$$U_t = \Omega(u, u_x, uu_x, u^2 u_x, u_{xx}, ...) \cdot \vec{\xi}, \qquad (1)$$

where $U_t$ is the left-hand side (LHS) term which can be the 1st or 2nd-order derivative of time; $\Omega$ is the linear combination operator; and $\vec{\xi}$ is the coefficient vector. For PDE discovery, it aims to identify several terms with non-zero coefficients from unlimited combinations of $u$ and its spatial derivatives.

The overview of the PIC criterion is provided in Fig. 1. An artificial neural network (ANN) is utilized here to construct a surrogate model to generate smoothed meta-data and calculate derivatives. Considering that solving Eq. (1) is an NP-hard problem with unlimited combinations, the generalized genetic algorithm optimization (illustrated in Fig. 1b) is employed to select an optimized preliminary library $\Phi_{opt}$ from unlimited combinations and convert the problem to a finite-dimensional problem (Fig. 1a). For the optimized preliminary library with a few terms, the possibilities of the combinations are countable, and thus it is feasible to measure the PIC of each combination $\Phi_j$ and select the most proper PDE. Calculating PIC involves two measurements: redundancy loss (r-loss) measuring the parsimony; and physical loss (p-loss) measuring the precision. Lejarza and Baldea[20] proved that the coefficients of redundant terms are unstable in the moving horizon, thus resulting in a high coefficient of variation (cv). Therefore, the r-loss is calculated by the average cv of PDE coefficients for each combination (Fig. 1c). Considering that solving potential PDEs to obtain precision is nearly infeasible, an alternative way utilizing PINN is adopted here. For the PINN, a correct physical constraint will improve the performance (and vice versa), which indicates that if the potential PDE is inconsistent with the observed data, the outcome of PINN will deviate from the outcome of ANN with the wrong physical constraint. Therefore, p-loss is measured by the deviation between the output of PINN and ANN (Fig. 1d). Finally, the PIC is calculated by the multiplication of r-loss and p-loss, and a smaller PIC indicates a better discovered PDE (Fig. 1e). The details of the PIC are provided in the Methods section. After the ultimate structure of PDE is determined by PIC, the coefficients can be further optimized by the PINN to obtain better accuracy[18].

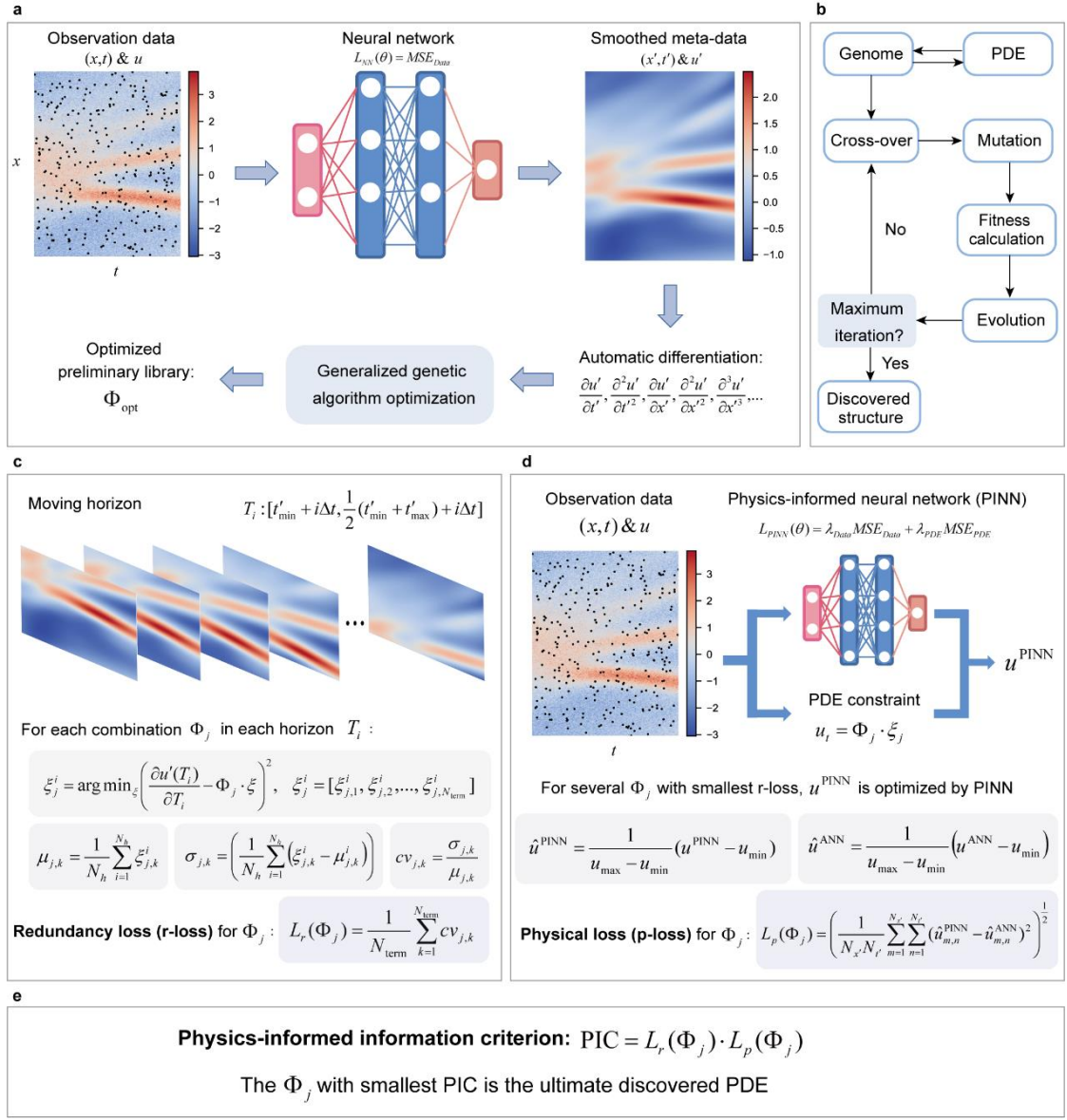

**Fig. 1. Overview of the PIC. a,** The flow chart of training the artificial neural network, calculating derivatives via automatic differentiation, and obtaining the optimized preliminary library. **b,** The flow chart of the generalized genetic algorithm optimization. **c,** The process of calculating redundancy loss for each possible combination by the moving horizon technique. Here, $N_h$ is the number of moving horizons, and $N_{term}$ is the number of terms in the combination $\Phi_j$. **d,** The process of calculating the physical loss by training PINN with potential PDEs as the physical constraints. **e,** The definition of the PIC.

**Discovery of canonical PDEs via PIC.** In this section, we utilize the PIC to identify several canonical PDEs, including the Korteweg-De Vries (KdV) equation, the Burgers equation, the convection-diffusion equation, the Chaffee-infante equation, the Allen-Chan equation, the wave equation, and the Klein-Gordon (KG) equation. These PDEs are originated from different physical fields, such as fluid mechanics and quantum mechanics. Here, the datasets are obtained from numerical solution, which is illustrated in Extended Data Fig. 1. In this work, the Gaussian noise is

added as:

$$\tilde{u} = \varepsilon \cdot std(u) \cdot N(0,1) + u, \qquad (2)$$

where $u$ are the clean data; $\tilde{u}$ are the noisy data; $N(0,1)$ is the standard normal distribution; and $\varepsilon$ is the noise level. In order to explore the robustness of the PIC to high noise, we increase the noise level by 25% until the PDE form fails to be discovered. We consider two types of activation functions, including the Sin function[21] and the Rational function[22], and select the better one when identifying each PDE. Additional details about these two activation functions are provided in the Supplementary Information. The results are displayed in Table 1. From the table, it is evident that the PIC method achieves outstanding performance when dealing with high levels of noise. Notably, in some cases, the PIC method is robust to extremely high noise (200% for the convection-diffusion equation, the KG equation, and 175% for the wave equation). For all seven canonical PDEs, state-of-the-art robustness to data noise is obtained with high accuracy of discovered coefficients.

**Table 1. The result of the PIC when identifying seven canonical PDEs from data with high levels of noise.** The noise level is the maximum noise where the correct PDE form can be discovered, and the activation function is the better one when identifying each PDE. The utilized data sizes are 10,000 for all cases.

| Equation name | Equation form | Activation function | Noise level | Discovered equation |
|---|---|---|---|---|
| KdV equation | $u_t = -uu_x - 0.0025u_{xxx}$ | Sin | 100% | $u_t = -1.043uu_x - 0.0026u_{xxx}$ |
| Burgers equation | $u_t = -uu_x + 0.1u_{xx}$ | Rational | 75% | $u_t = -0.960uu_x + 0.078u_{xx}$ |
| Convection diffusion equation | $u_t = -u_x + 0.25u_{xx}$ | Rational | 200% | $u_t = -1.02u_x + 0.251u_{xx}$ |
| Chaffee-Infante equation | $u_t = u_{xx} - u + u^3$ | Rational | 50% | $u_t = 1.108u_{xx} - 1.312u + 1.117u^3$ |
| Allen-Cahn equation | $u_t = 0.003u_{xx} + u - u^3$ | Rational | 50% | $u_t = 0.003u_{xx} + 0.970u - 1.007u^3$ |
| Wave equation | $u_{tt} = u_{xx}$ | Sin | 175% | $u_{tt} = 0.976u_{xx}$ |
| KG equation | $u_{tt} = 0.5u_{xx} - 5u$ | Rational | 200% | $u_{tt} = 0.465u_{xx} - 5.55u$ |

To further investigate the robustness of the PIC, the maximum noise where the PIC can discover the correct PDE form with different data sizes is provided in Fig. 2a. Here, the convection-diffusion equation is taken as an example. From the figure, the robustness to data noise increases with data sizes, and the performance is satisfactory with sparse data since it is still robust to 25% noise with merely 100 data. Meanwhile, the relative coefficient error is utilized to measure the difference between the discovered PDE and the true PDE, which is defined as follows:

$$e_{coef} = \frac{1}{N_{term}} \sum_{i=1}^{N_{term}} \left| \frac{\xi_i - \xi_i^{true}}{\xi_i^{true}} \right|, \qquad (3)$$

where $N_{term}$ is the number of PDE terms; and $\xi_i$ and $\xi_i^{true}$ are the coefficient of the discovered and the true PDE term, respectively. The relative coefficient error for different noise levels when identifying the KG equation is illustrated in Fig. 2b. Here, we conducted 10 independent experiments with different random seeds for each noise level to present a statistical result. It is evident that the relative error has an increasing trend with the noise level, but remains relatively low with high levels of noise (only 4.21% median relative error for 200% noise).

Another experiment is conducted to examine the robustness of the hyper-parameters. In most PDE discovery methods, the core hyper-parameters that control the regularization or threshold significantly impact the discovered PDE since they decide the balance between parsimony and precision. The hyper-parameters are usually different in different situations and have to be fine-tuned to obtain the correct PDE[23], which is unrealistic for practical applications. In the PIC, the most influential hyper-parameter is the $l_0$-penalty that determines the size of the preliminary library. We take the KG equation with 100% noise as an example, and the results with different magnitudes of $l_0$-penalty are provided in Fig. 2c. The LHS term $U_t$ is supposed to be unknown here, and the best combinations obtained with the hypothesis that the LHS term is $u_t$ or $u_{tt}$ are shown in the figure. It is seen that the size of the preliminary library increases when $l_0$-penalty decreases. However, the ultimate discovered PDE is stable for different magnitudes of $l_0$-penalty.

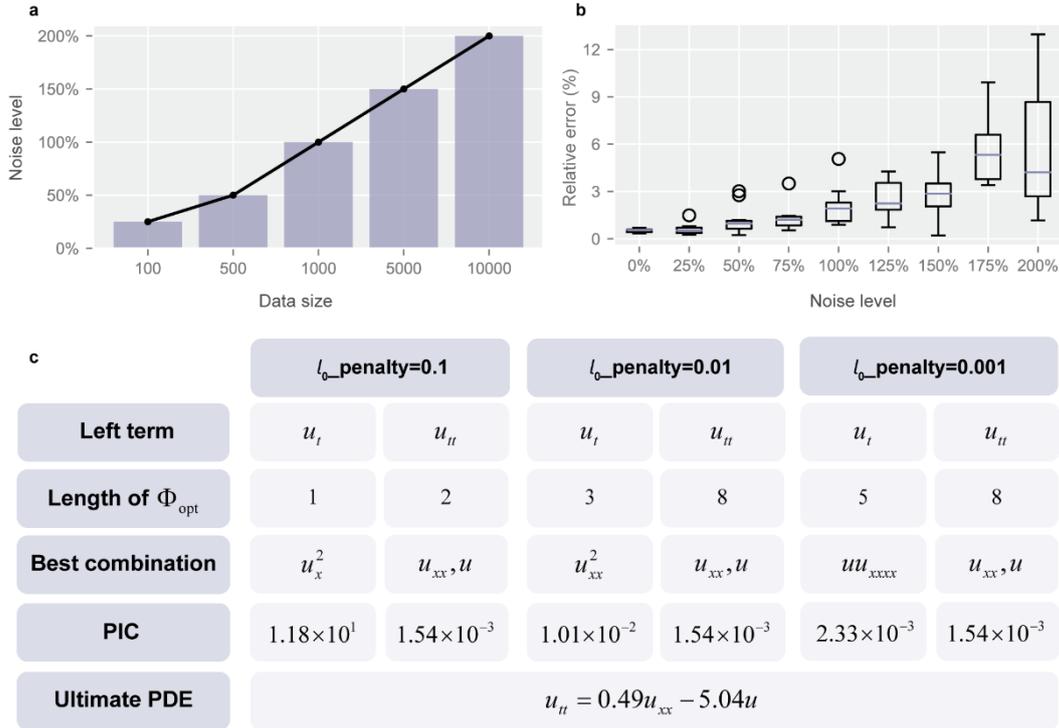

Fig. 2. The robustness of PIC criterion to sparse data, high levels of noise, and hyper-parameters. **a**, The maximum noise where the PIC criterion can discover the correct PDE form in different data sizes when identifying the convection-diffusion equation. **b**, The boxplot of relative coefficient error for the ultimately discovered PDE from the data with different noise levels when identifying the KG equation with 10 different random seeds. The pink lines are the medians, and the black dots are the outliers **c,** The PIC and ultimate PDE when discovering the KG equation with 100% noise and different hyper-parameters, i.e., $l_0$_penalty, which controls the length of the preliminary potential terms $\Phi_{opt}$. Here, the LHS term $U_t$ is unknown, and the best combination

discovered of the $u_t$ and $u_{tt}$ as the LHS term is compared to select the best one with a smaller PIC.

The high robustness and accuracy of the PIC are attributable to several aspects. Firstly, the neural network can smooth the data noise to some extent. The generalized genetic algorithm can provide a complete preliminary library since it has been proven to discover the redundant compensation terms to guarantee the inclusion of the correct dominant terms as far as possible in cases of high noise[18]. This issue is also confirmed in the Supplementary Information by experiments. Secondly, the PIC criterion takes both parsimony and precision into account, which can select the optimal PDE from numerous combinations from the preliminary library. Finally, the high accuracy of the ultimate PDEs is ascribed to the optimization of PINN. If the correct structure is added as the physical constraints, an optimization loop will be constructed to improve the accuracy. In other words, a more accurate PDE constraint leads to a neural network that fits the data better so that the coefficients are optimized to be more precise, which makes the PDE constraint more accurate[18].

**Comparison with other methods and criteria.** In this section, the PIC is compared with existing commonly-used methods and criteria to demonstrate its superiority further. First, the PIC is compared with the AIC and the BIC criterion[24], which are often employed to measure the complexity and accuracy of the estimated model. We adopt them to identify optimal structures from the preliminary potential library to better demonstrate the difference between these information criteria. Here, the KdV equation with 100% noise is taken as an example, and the optimized preliminary library has three terms that make up eight combinations. The top-three discovered structures with these information criteria are shown in Fig. 3a. It can be seen that AIC and BIC have difficulty distinguishing the redundant terms when faced with high levels of noise. At the same time, the PIC can successfully identify the best structure since the PIC of the true structure is apparently smaller. Meanwhile, the coefficients obtained by PIC are more accurate because of the PINN utilized in PIC.

Afterwards, the PIC criterion is compared with Lasso (utilizing $L_1$ penalty) and STRidge (using $L_2$ penalty and $L_0$ penalty). Considering that both Lasso and STRidge require a complete candidate library, we utilize a basic candidate library with 10 terms:

$$\Phi = [u, u_x, u_{xx}, u_{xxx}, uu_x, uu_{xx}, uu_{xxx}, u^2 u_x, u^2 u_{xx}, u^2 u_{xxx}]. \tag{4}$$

Here, the derivatives are also calculated from the automatic differentiation. The KdV equation with 100% noise is taken as an example here. The discovered structures with different hyper-parameters are demonstrated in Fig. 3b (Lasso) and Fig. 3c (STRidge). For the Lasso method, a larger alpha ($L_1$ penalty) corresponds to a more parsimonious structure, but the true structure fails to be discovered with all magnitudes of alpha faced with highly noisy data. The STRidge method performs better since the correct dominant terms are contained in the discovered structure with different $L_0$ penalties. However, the discovered structure has redundant terms with a small $L_0$ penalty, but lacks correct terms with a large $L_0$ penalty. In contrast, the result of PIC is both stable and accurate.

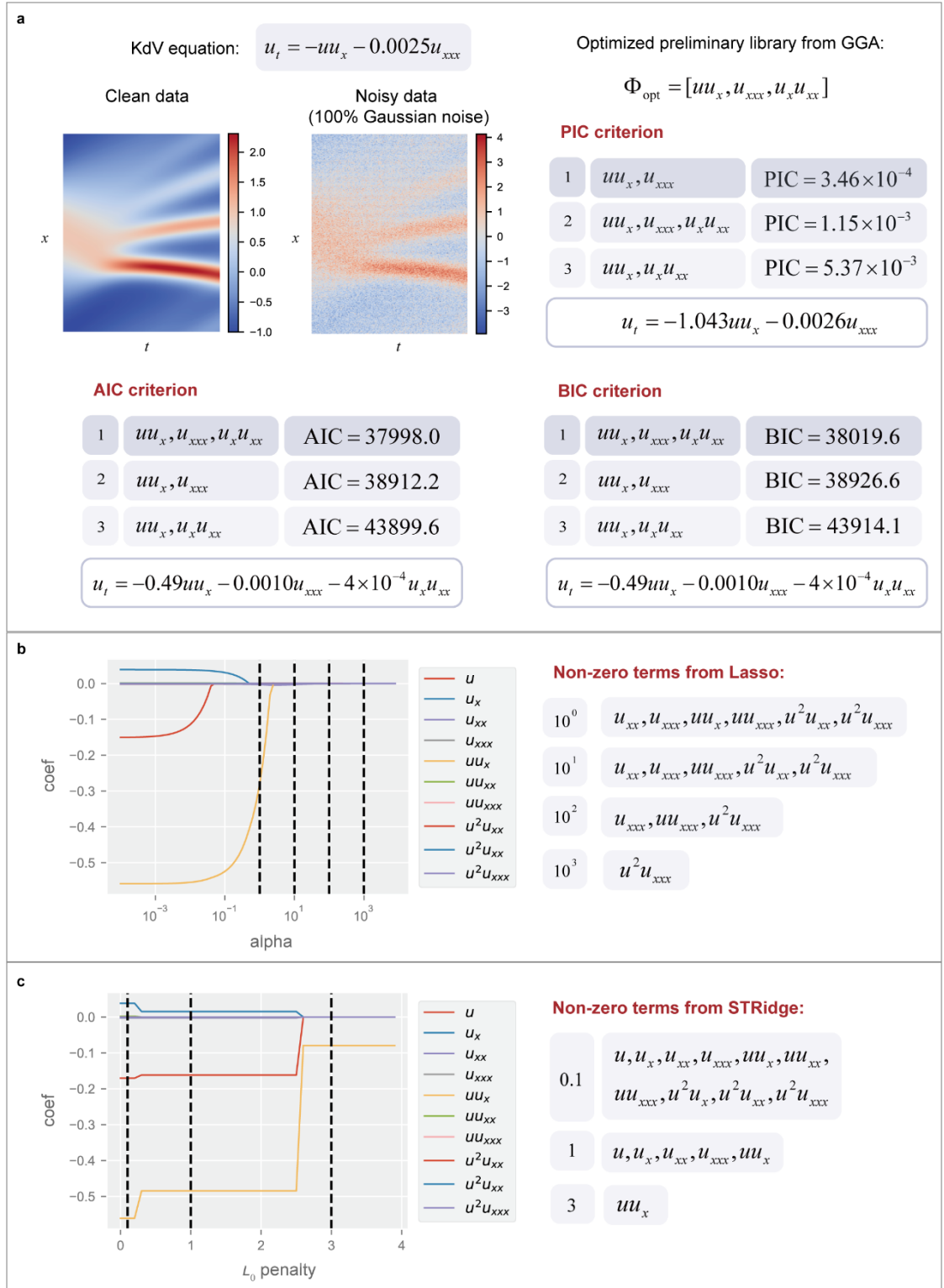

**Fig. 3. Comparison between existing criteria and methods. a**, The top-three identified structures and ultimately discovered PDEs by PIC (upper right), AIC (bottom left), and BIC (bottom right). **b,** The non-zero terms discovered from the Lasso with different alpha, which controls the magnitude of $L_1$ regulation. **c,** The non-zero terms discovered from the STRidge with different $L_0$ penalties. The above results are obtained when discovering the KdV equation with 100% noise. The black dashed line corresponds to the situation with different hyper-parameters.

**Extendibility and practical applications.** In this section, the extendibility of the PIC to broader situations is investigated. The canonical PDEs examined in this work are all 1D cases, and the extension to higher dimensional conditions is straightforward and easy, details of which are provided in the Methods section. Here, we adopt the PIC criterion to discover the 2D Burgers equation with 200,000 discrete data and different noise levels, the results of which are presented in Table 2. The true PDE form is written as:

$$u_t = -(uu_x + uu_y) + 0.01(u_{xx} + u_{yy}). \tag{5}$$

From the table, it is discovered that the PIC criterion is robust up to 75% noise, and the discovered coefficient is accurate. The best results in existing works were achieved by Zhang and Liu[15], who utilize 515,100 grid data, and their method is robust to 40% Gaussian noise. This indicates that the PIC can also discover high-dimensional PDEs with high accuracy and robustness.

We then investigate the extendibility of PIC to discover parametric PDEs with variant coefficients, which is a challenge for most existing methods. The parametric convection-diffusion is taken as an example, the form of which is written as:

$$u_t = -(1 + 0.25\sin(x))u_x + u_{xx}. \tag{6}$$

The results are shown in Extended Data Table 1. It is discovered that PIC is robust to 50% noise, which is a great improvement compared with the 25% noise obtained by Xu et al.[25] with the same condition. Furthermore, faced with 75% noise, although the coefficients' form has a deviation, the correct structure is still discovered by PIC. Additional details about the extension to parametric PDEs can be found in the Methods section and Supplementary Information.

**Table 2. The discovered PDE and corresponding PIC from data with different noise levels when identifying the 2D Burgers equation.**

| Noise level | Discovered equation | PIC |
|---|---|---|
| 0% | $u_t = -0.998(uu_x + uu_y) + 0.0093(u_{xx} + u_{yy})$ | $1.60 \times 10^{-5}$ |
| 25% | $u_t = -1.006(uu_x + uu_y) + 0.0095(u_{xx} + u_{yy})$ | $7.08 \times 10^{-6}$ |
| 50% | $u_t = -1.047(uu_x + uu_y) + 0.0093(u_{xx} + u_{yy})$ | $7.52 \times 10^{-6}$ |
| 75% | $u_t = -1.056(uu_x + uu_y) + 0.0092(u_{xx} + u_{yy})$ | $9.83 \times 10^{-6}$ |
| 100% | $u_t = 0.133(uu_{xx} + uu_{yy}) - 0.040(u_x + u_y)$ | $4.82 \times 10^{-5}$ |

Finally, the PIC is employed to discover unrevealed PDEs from a practical physical scene. In this work, we focus on a practical physical case of proppant transport, the process of which is illustrated in Fig. 4a. It describes the relative motion of two fluids with different densities and viscosities in a fracture or channel. The essence of the problem is similar to the viscous gravity current, which is an important natural phenomenon in geophysics. The asymptotic behavior of the developed current has been studied deeply in previous works[26], while the macroscale governing equation of the early-time behavior is hard to be derived theoretically. Here, we will investigate the preasymptotic behavior at a macroscopic scale because of its importance in depicting proppant transport in fractures during the rapid process of hydraulic fracturing. As shown in Fig. 4a, the boundary between the two fluids is vertical under initial conditions, and the height of the boundary $h(x)$ develops with time because of the influence of gravity. Here, we consider two situations, including $\mu_1 = \mu_2$ and $\mu_1 \gg \mu_2$, where $\mu_1$ and $\mu_2$ are viscosities of fluid 1 and 2, respectively. The dataset $h(x,t)$ is extracted from high-resolution microscopic simulation results and is standardized to obtain a general result. It is worth noting that the dataset is obtained from the numerical simulation at a microscopic scale[27], instead of solving known macroscale PDEs, because the theoretical governing equations at the macroscopic scale for this process have not been revealed previously. Here, the integration technique is adopted to facilitate the discovery of the macroscale governing equations, which is detailed in the Methods section. The ultimate discovered PDEs by PIC are provided in Fig. 4b ($\mu_1 = \mu_2$) and Fig. 4c ($\mu_1 \gg \mu_2$). The figure shows that the posterior data obtained from the solution of the discovered PDE are consistent with the observation data with low error. Here, the relative error is defined as follows:

$$e = \left( \frac{\sum_{j=1}^{N_t} \sum_{i=1}^{N_x} \left| h(x_i, t_j) - h'(x_i, t_j) \right|^2}{\sum_{j=1}^{N_t} \sum_{i=1}^{N_x} \left| h(x_i, t_j) \right|^2} \right)^{\frac{1}{2}} \times 100\%, \tag{7}$$

where $h(x_i, t_j)$ are the observation data; $h'(x_i, t_j)$ is the solution of the discovered PDE; and $N_x$ and $N_t$ are the number of $x$ and $t$, respectively. The relative error for case 1 and case 2 is 1.42% and 2.10%, respectively, which indicates that the discovered PDE can fit the observed data very well. Furthermore, the discovered PDE is parsimonious and interpretable. For case 1, the discovered PDE can be rewritten as:

$$\frac{\partial h^*}{\partial t^*} = \frac{\partial}{\partial x^*} \left( 0.776 h^* (1 - 1.01 h^*) \frac{\partial h^*}{\partial x^*} \right), \tag{8}$$

which can be seen as a parametric diffusion equation where the variant coefficients are related to the height $h^*$, and the equation can be transferred into:

$$\frac{\partial h^*}{\partial t^*} = \frac{\partial}{\partial x^*} \left( \alpha(h^*) \frac{\partial h^*}{\partial x^*} \right), \tag{9}$$

where $\alpha(h^*) = 0.776 h^* (1 - 1.01 h^*)$. In essence, the variant coefficients characterize the diffusion trend at different heights. Meanwhile, it is surprising that the discovered PDE approximately satisfies the underlying symmetry $\alpha(h^*) = \alpha(1 - h^*)$, which is consistent with the property of the physical process. Here, $h^*$ and $1-h^*$ represent the height of fluid 1 and fluid 2, respectively, and if $\mu_1 = \mu_2$, they are equivalent and symmetry exists. Similarly, for case 2, the discovered PDE is also a parametric diffusion equation while the variant coefficient $\alpha(h^*) = 0.751 h^*$. The result can be explained physically because if $\mu_1 \gg \mu_2$, the dynamic pressure of fluid 2 is much less than that of fluid 1, which means that fluid 1, with much larger viscosity, plays a leading role. Therefore, the variant coefficient is directly related to $h^*$ representing the height of fluid 1.

In general, the governing equations for the proppant transport process are discovered by PIC, which were previously unrevealed. The discovered PDE can describe the physical process well and is consistent with the observation data from numerical simulation. Meanwhile, the discovered PDE is parsimonious and thus easy to be solved and explained, which advances the understanding and description of the preasymptotic behavior of the viscous gravity current. In addition, it also reveals underlying symmetries behind the physical process. The governing equations discovered by PIC can facilitate the understanding and solution of the physical process because solving PDEs is much faster than high-resolution microscopic simulation. It is worth noting that the proppant transport process has also been studied in our earlier work[27] with the existing PDE discovery method, but complex redundant terms in the result make the discovered PDE hard to be solved and explained, and it is highly challenging for symmetries to be revealed.

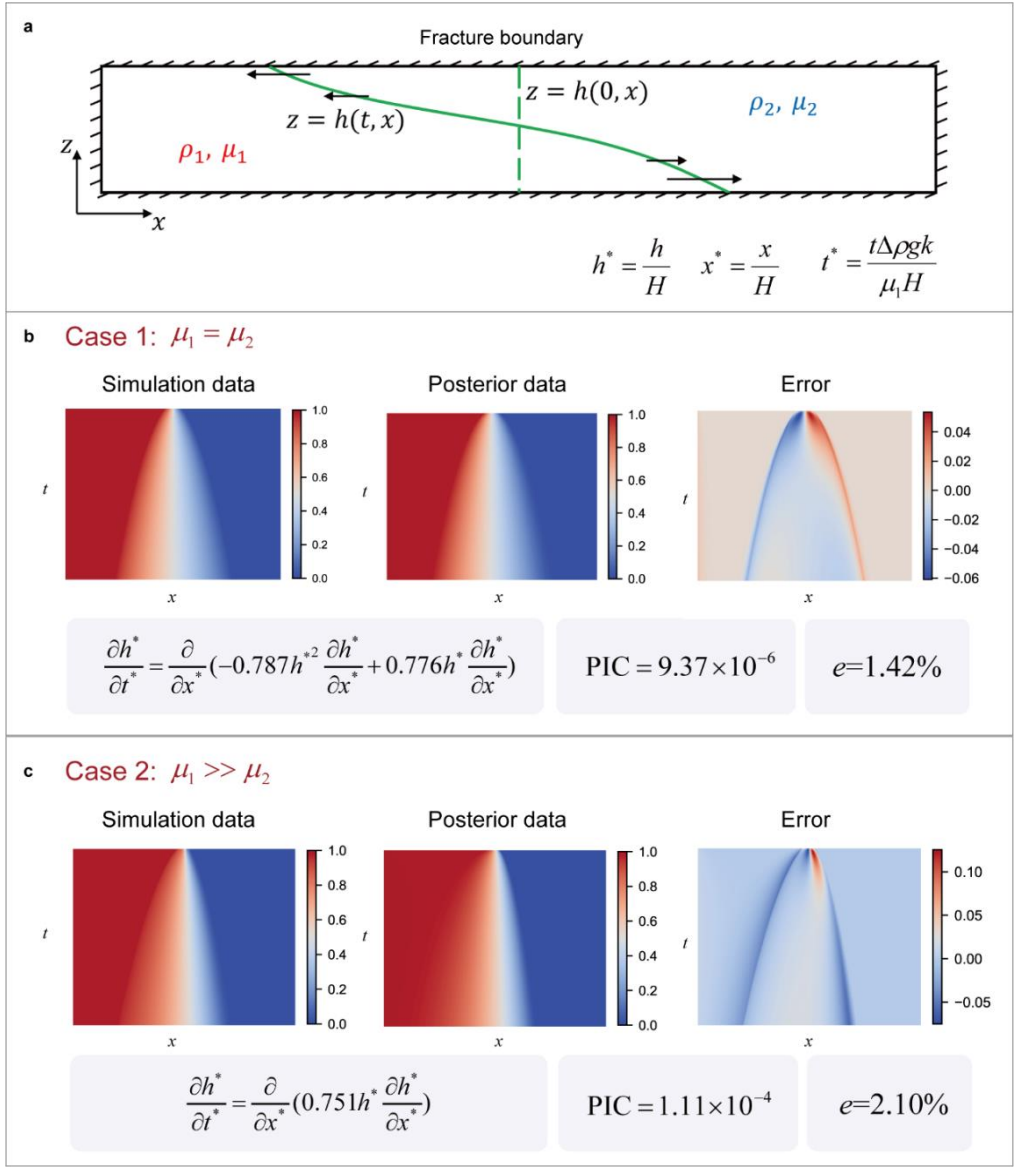

**Fig. 4. Practical applications of PIC to discover unrevealed governing equations in proppant transport. a**, Simplified diagram of the physical process for proppant transport and dimensionless variable. The observation data are obtained from high-resolution microscopic simulation results. **b,** The simulation data, posterior data solved from the discovered PDE, and the error in the case of $\mu_1 = \mu_2$. **c,** The simulation data, posterior data solved from the discovered PDE, and the error in the case of $\mu_1 \gg \mu_2$. $e$ is the relative error.

**Discussion**

PDE discovery essentially constitutes a balance of precision and parsimony, which is controlled by regularization and threshold in previous literature. However, in terms of practical applications, problems will emerge in which it is difficult to determine the most proper PDE without the referenced PDE since the results are usually different with different hyper-parameters characterizing regularization. Therefore, in this work, an innovative criterion, the physics-informed information criterion (PIC), is proposed to determine the optimal PDE for both proof-of-concept cases and practical applications. State-of-the-art outcomes are obtained for canonical PDEs from different physical scenes, proving PIC's ability to deal with sparse and highly noisy data. Meanwhile, the stability of PIC to hyper-parameters makes it more suitable for broader tasks. Finally, the PIC is employed in a specific physical case of proppant transport, which is essentially a viscous gravity current in the fracture. For this physical process, the unknown macroscale governing equation of the preasymptotic behavior is successfully discovered by PIC from microscale simulation data without any other prior information. It is found that the discovered governing equation has good precision and interpretability, and even satisfies potential symmetries, which facilitates the understanding and description of the preasymptotic behavior of the viscous gravity current. However, the computational cost of the PIC is currently higher than other methods, especially sparse regression-based methods, due to the need to train multiple neural networks. Meanwhile, the possible combinations increase with the size of the preliminary potential library with exponential growth, which is greatly time-intensive to traverse all possibilities when the size of the preliminary potential library is large. Nevertheless, the computation is affordable (usually taking 7~20 min in this work), although on-going efforts attempt to optimize the PDE discovery process. In summary, the proposition of PIC has made further satisfactory progress in practical applications of PDE discovery, and will facilitate the discovery of unrevealed governing equations in broader physical scenes.

**Methods**

**The generalized genetic algorithm optimization.** In this work, a generalized genetic algorithm is utilized to provide the preliminary potential library, which originated from Xu et al.[18]. Here, we briefly introduce the process of the algorithm. As illustrated in Fig. 1b, the generalized genetic algorithm consists of several important steps, including translation, cross-over, mutation, fitness calculation, and evolution.

The PDE is digitized into genomes composed of inner terms and corresponding derivatives order in the translation step. For inner terms, the number refers to the corresponding derivative order, which is called basic genes. For example, 1 refers to $u_x$ for the right-hand side (RHS) term or $u_t$ for the left-hand side (LHS) term, and 2 refers to $u_{xx}$ or $u_{tt}$, similarly. The inner term is composed of the multiplication of basic genes. For example, (1,2) refers to the inner term $u_x u_{xx}$. Then, the term (or gene module) is constructed with the inner term and corresponding derivatives order. For example, [(1,2),1] refers to $\frac{\partial(u_x u_{xx})}{\partial x}$. Finally, several terms consist of the PDE by addition. For instance, {[(1,2),1], [(0,0),2]} refers to $\frac{\partial(u_x u_{xx})}{\partial x} + \frac{\partial^2(u^2)}{\partial x^2}$. Similarly, the LHS terms can be translated into genomes. With this digitization principle, each PDE corresponds to a specific genome, which paves the way for the subsequent process. The genetic algorithm is generalized because it can represent broader forms of terms, such as compound forms. For the initial generation, a number of genomes are randomly generated.

In the cross-over step, children are produced by swapping certain terms (or gene modules) in two parent genomes. In the mutation step, the children will mutate to generate new possibilities. There are three mutation ways, including: delete-module mutation, in which a randomly chosen module is deleted; add-module mutation, in which a randomly generated module is added; and basic gene mutation, in which a certain basic gene is replaced by a new randomly generated one. These two steps are crucial to the performance of the generalized genetic algorithm because unlimited combinations can be generated in these steps, which expand the search scope.

In the fitness calculation step, each genome's fitness is calculated to measure the quality of genomes, which is defined as follows:

$$F = MSE + \varepsilon \cdot N_{term}, \tag{10}$$

with

$$MSE = \frac{1}{N_x N_t} \sum_{j=1}^{N_t} \sum_{i=1}^{N_x} \left| U_L(x_i, t_j) - \vec{\xi} U_R(x_i, t_j) \right|^2, \tag{11}$$

where $F$ denotes the fitness, which consists of $MSE$ part and $l_0$ penalty; $MSE$ is calculated according to Eq. (11), where $U_L(x_i, t_j)$ is the value of the LHS term with the size of $1 \times 1$; $U_R(x_i, t_j)$ is the value of RHS terms translated from the genome, which is a $N_{term} \times 1$ vector; $N_{term}$ is the number of PDE terms; $N_x$ and $N_t$ are the number of $x$ and $t$ of the meta-data, respectively; and $\vec{\xi}$ with the size of $1 \times N_{term}$ is the coefficient calculated by the least squares solution of $U_L - \vec{\xi} U_R = 0$, which can be solved by the SVD method. Here, a smaller fitness indicates a better genome.

After the fitness has been calculated, half of the genomes with smaller fitness are reserved, and the others are replaced by new randomly generated genomes, which consist of the next generation. The evolution process will continue until the maximum iteration is achieved, and the optimal structure is the best genome with the smallest fitness in the last generation. It is worth noting that the compound form in the optimal structures will be decomposed and merged into the simplest form, which consists of the preliminary potential library.

Compared with other optimization methods, the generalized genetic algorithm possesses the advantage of broader representation and more robustness to data noise. It has been proven that the generalized genetic algorithm is able to discover the redundant compensation terms to guarantee the inclusion of the correct dominant terms[18], which is also demonstrated by an experiment provided in the Supplementary Information. The inclusion of the correct dominant terms is important for constructing a complete preliminary library from unlimited combinations, which paves the way for the PIC.

**The moving horizon technique and calculation of redundancy loss (*r*-loss).** The moving horizon is usually utilized for state and parameter estimation[28]. Lejarza and Baldea[20] first utilized this technique to facilitate model selection in ODE discovery. In this work, the moving horizon technique is employed to calculate *r*-loss to measure the PDE's parsimony. As illustrated in Fig. 1c, the smoothed meta-data are divided into $N_h$ overlapping horizons $T_i$, which is defined as $[t'_{min}+i\Delta t, \frac{1}{2}(t'_{min}+t'_{max})+i\Delta t]$, where $t'_{min}$ and $t'_{max}$ are the minimum and maximum of the time domain of the meta-data $t'$, respectively; $i=1,2,...,N_h$; and $\Delta t$ is the length of horizons. The meta-data in horizons $T_i$ are generated from the neural network. For a given combination (i.e., PDE structure) $\Phi_j$, the optimal coefficients for each term in horizons $T_i$, $\xi_j^i$, can be calculated by solving $U_L^{i,j} - \xi_j^i \cdot U_R^{i,j} = 0$, where $U_L^{i,j}$ and $U_R^{i,j}$ are the values of the LHS term and RHS terms for $\Phi_j$ in $T_i$, respectively. Therefore, for each term in $\Phi_j$, $N_h$ different coefficients are obtained, and the coefficient of variation (*cv*) can be calculated as:

$$cv_{j,k} = \frac{\sigma_{j,k}}{\mu_{j,k}}, \quad (12)$$

where $\sigma_{j,k}$ and $\mu_{j,k}$ are the standard deviation and mean of the $N_h$ different coefficients for the $k^{th}$ term in $\Phi_j$, respectively. The *r*-loss of $\Phi_j$ is calculated by the mean *cv* of all terms:

$$L_r(\Phi_j) = \frac{1}{N_{term}} \sum_{k=1}^{N_{term}} cv_{j,k}, \quad (13)$$

where $L_r(\Phi_j)$ is the *r*-loss for the combination $\Phi_j$; and $N_{term}$ is the number of terms. With the moving horizon, the *r*-loss can measure the parsimony well because the redundant terms function as the compensation for the error caused by noise, which is different in different horizons, leading to a high *cv*. In contrast, the correct dominant terms are stable with smaller *cv*. More details and discussions are provided in the Supplementary Information.

**The physics-informed neural network and calculation of physical loss (*p*-loss).** The physics-informed neural network (PINN) was first proposed by Raissi et al.[16] to construct surrogate models and improve prediction performance. Compared with common neural networks, the PINN adds physical constraints to the loss function of the neural network to make the prediction fit the prior physical knowledge well. As illustrated in Fig. 1d, in this work, the PINN is utilized to measure the precision of the discovered PDE without solving the PDE numerically. The construction of the PINN is the same as ANN, which consists of one input layer, several hidden layers and one output layer, while the loss of physical constraint is added to the loss function:

$$L_{PINN}(\theta) = \lambda_{Data}MSE_{Data} + \lambda_{PDE}MSE_{PDE},$$

$$MSE_{Data} = \frac{1}{N_x N_t}\sum_{i=1}^{N_x}\sum_{j=1}^{N_t}(u(x_i,t_j) - PINN(x_i,t_j;\theta)), \quad (14)$$

$$MSE_{PDE} = \frac{1}{N'_x N'_t}\sum_{i=1}^{N'_x}\sum_{j=1}^{N'_t}(U'_L(x'_i,t'_j) - \xi' \cdot U'_R(x'_i,t'_j))^2.$$

where $L_{PINN}(\theta)$ is the loss function of PINN, which consists of data loss ($MSE_{Data}$) and PDE loss ($MSE_{PDE}$); and $\theta$ is the parameters of PINN, including the weights and bias. The data loss is the mean squared error (*MSE*) of the difference between observed data and predicted data by PINN. The PDE loss is the *MSE* of the difference between the LHS term $U'_L$ and the RHS term $\xi' \cdot U'_R$ of the PDE. Here, the PDE loss is calculated on the meta-data $(x'_i, t'_j)$ generated from the neural network. It is worth noting that the coefficients of the PDE $\xi'$ are calculated by solving $U'_L - \xi' \cdot U'_R = 0$ in each training epoch. In this work, the PINN is trained on the basis of pre-trained ANN in Fig. 1a, and the input is the observation data $(x_i, t_j)$, as well. The output of PINN on the meta-data is $u^{PINN}$, while the output of the pre-trained ANN on the meta-data is $u^{ANN}$. To calculate *p*-loss, we standardize the output by:

$$\hat{u}^{PINN} = \frac{1}{u_{max} - u_{min}}(u^{PINN} - u_{min}), \quad (15)$$

$$\hat{u}^{ANN} = \frac{1}{u_{max} - u_{min}}(u^{ANN} - u_{min}),$$

where $u_{max}$ and $u_{min}$ are the maximum and minimum of the observation data, respectively. Then, the *p*-loss is calculated by the root mean squared error (RMSE) between the standardized output of PINN and ANN, which is denoted as follows:

$$L_p(\Phi_j) = \left(\frac{1}{N_{x'}N_{t'}}\sum_{m=1}^{N_{x'}}\sum_{n=1}^{N_{t'}}(\hat{u}^{PINN}_{m,n} - \hat{u}^{ANN}_{m,n})^2\right)^{\frac{1}{2}}. \quad (16)$$

In this work, we use the PINN to measure the precision of the discovered PDE. It is based on the fact that the physical constraint has an influential impact on the prediction. If the physical constraint is parallel with the data, the prediction will be improved. However, if the physical constraint and the data are inconsistent, the training will be affected significantly, and the result will derive a lot from the output of ANN. Since ANN's output is relatively accurate, the *p*-loss will be very small if the PDE can describe the data. Therefore, the *p*-loss can measure the precision efficiently without solving PDEs. Additional details about *p*-loss are provided in the Supplementary Information.

**The physics-informed criterion (PIC).** As illustrated in Fig. 1e, the PIC is defined by the multiplication of the calculated *p*-loss and *r*-loss, which is written as:

$$\text{PIC} = L_r(\Phi_j) \cdot L_p(\Phi_j). \tag{17}$$

Here, a small PIC means that the PDE has both better parsimony and precision. The whole PIC algorithm is detailed below. In this work, an artificial neural network (ANN) is pre-trained to construct a surrogate model to generate smoothed meta-data and calculate derivatives. Then, the generalized genetic algorithm optimization is employed to obtain an optimal structure that forms the preliminary potential library $\Phi_{opt}$. For the preliminary potential library with countable terms, the *r*-loss of all possible combinations is calculated. Here, if the preliminary library has $N_{opt}$ terms, the total number of combinations is $2^{N_{opt}}$. The potential library is unlimited in common practice, which poses an NP-hard problem. This is why we utilize the generalized genetic algorithm in this work to obtain a countable preliminary library. It is worth noting that the size of the preliminary library is usually smaller than 10, and the computational cost is reasonable because the calculation process of *r*-loss is quick for each combination. After the *r*-loss has been calculated, we select top $N_b$ combinations with smaller *r*-loss and further calculate *p*-loss. Considering that the training of PINN is time-consuming, it is infeasible to calculate *p*-loss of all combinations. Afterwards, the PIC of the $N_b$ combinations is calculated, and the ultimate discovered PDE is the combination with the smallest PIC. Finally, the coefficient of the discovered PDE is further optimized by the PINN.

**The canonical PDEs in this work.** In this work, seven canonical PDEs from different physical fields are investigated to examine the robustness and accuracy of the proposed PIC for proof-of-concept. Here, we will briefly introduce these PDEs, and the visualization of their dataset is provided in Extended Data Fig. 1.

1. The Korteweg-De Vries (KdV) equation, which is utilized to describe the evolution of one-dimensional shallow-water waves, is written as:

$$u_t = -uu_x - 0.0025 u_{xxx}. \tag{18}$$

The dataset is grid data of 512 spatial observation points in the domain $x \in [-1,1)$ and 201 temporal observation points in the domain $t \in [0,1]$, and thus the data size is 102,912.

2. The Allen-Chan equation, which is a nonlinear reaction-diffusion equation that describes the phase separation of multi-component metal alloys[29], is written as:

$$u_t = 0.003 u_{xx} + u - u^3. \tag{19}$$

The dataset is grid data of 256 spatial observation points in the domain $x \in (-1,1)$ and 201 temporal observation points in the domain $t \in (0,10]$, and thus the data size is 51,456.

3. The wave equation, which is a commonly-used PDE to describe the vibration and fluctuation phenomenon, is written as:

$$u_{tt} = u_{xx}. \tag{20}$$

The dataset is grid data of 161 spatial observation points in the domain $x \in [0, \pi]$ and 321 temporal observation points in the domain $t \in [0, 2\pi]$, and thus the data size is 51,681.

4. The convection-diffusion equation, which can be employed to describe the transport of substance in the fluid (e.g., contaminant transport), is written as:

$$u_t = -u_x + 0.25 u_{xx}. \tag{21}$$

The dataset is grid data of 256 spatial observation points in the domain $x \in [0, 2]$ and 100 temporal observation points in the domain $t \in [0, 1]$, and thus the data size is 25,600.

5. The Burgers equation, which has a wide application in many fields, including fluid mechanics, nonlinear acoustics, gas dynamics and traffic flow[30], is written as:

$$u_t = -u u_x + 0.1 u_{xx}. \tag{22}$$

The dataset is grid data of 256 spatial observation points in the domain $x \in (-8, 8)$ and 201 temporal observation points in the domain $t \in (0, 10]$, and thus the data size is 51,456.

6. The Klein-Gordon (KG) equation, which was first proposed by Oskar Klein and Walter Gordon in 1926 to describe the behavior of electrons in relativistic settings, is written as:

$$u_{tt} = 0.5 u_{xx} - 5u. \tag{23}$$

The dataset is grid data of 201 spatial observation points in the domain $x \in [-1, 1]$ and 201 temporal observation points in the domain $t \in [0, 3]$, and thus the data size is 40,401.

7. The Chaffee-infante equation, which is widely used in many fields, such as environmental science, fluid dynamics, high-energy physics and electronic science[11], is written as:

$$u_t = u_{xx} - u + u^3. \tag{24}$$

The dataset is grid data of 301 spatial observation points in the domain $x \in [0, 3]$ and 200 temporal observation points in the domain $t \in (0, 0.5)$, and thus the data size is 60,200.

**Extensibility to high-dimensional PDEs and parametric PDEs.** The PIC proposed in this work can be extended to different situations with only a slight adjustment. For high-dimensional PDEs, the derivatives of all dimensions are added to the basic genes of the generalized genetic algorithm, which leads to a preliminary library with high-dimensional terms. Here, we take the extension to 2D cases as an example, and Eq. (1) is converted into the following:

$$U_t = \Omega(u, u_x, u u_x, u_{xx}, u_y, u u_y, u_{yy} ...) \cdot \vec{\xi}, \tag{25}$$

where interactive terms, such as $u_x$ and $u_y$, are not considered. For high-dimensional problems,

considering that the search space is too large to be optimized, symmetry is always employed to facilitate the optimization and covert the problem into:

$$U_t = \Omega(u, \nabla u, (u \cdot \nabla)u, \nabla^2 u, ...) \cdot \vec{\xi}, \quad (26)$$

where the symmetrical terms are considered. For the generalized genetic algorithm, the basic genes are composed of the basic symmetrical terms, such as $u, \nabla u, \nabla^2 u$, etc. To maintain symmetry, we do not consider the compound form here. After these treatments, the PIC can be adapted to high-dimensional cases in a manner similar to the 1D case.

For parametric PDEs, the PIC can also assist to improve robustness based on the existing method. The core issue of identifying parametric PDEs is determining the PDE structure. In previous work[25], the optimal structure is identified by local windows and the voting principle, which may be ineffective when the noise level is high. Here, the PIC is adapted to improve the discovery of parametric PDEs. In each local window, a preliminary library can be obtained from the generalized genetic algorithm, which can be combined into a summarized library for the global domain. Then, the $r$-loss of all possible combinations of the summarized library is calculated, and the $p$-loss of several top combinations is calculated. The PDE loss is slightly different since the coefficient is variant, which is detailed in the Supplementary Information. Finally, the ultimate structure is discovered, and the variant coefficients are optimized in PINN. Furthermore, the curves of the variant coefficients can be fitted by the DLGA-PDE (coefficient) proposed by Xu et al.[25].

**Discovery of the governing equation of proppant transport by PIC.** In this work, the PIC is employed to discover the unrevealed governing equation of proppant transport. Here, we first briefly introduce the background of the physical scene. The proppant transport problem is originated from the hydraulic fracturing process where the fluid is injected into a crack, and the fluid pressure is the driving force for the fracture opening and propagation[31]. The proppant is used to prevent the fracture from closing once the well is depressurized, and the transport of proppant is affected by the original liquid in the crack. In this work, we consider a simplified situation to better reveal the process's essence. The dataset is generated from 2D high-resolution microscopic simulations with closed boundaries in a homogeneous porous medium or rectangle vertical fracture. The permeability and porosity are considered constant. In case 1 where $\mu_1 = \mu_2$, the dataset is grid data of 500 spatial observation points in the domain $x \in [0, 500)$ and 500 temporal observation points in the domain $t \in [0, 100)$, and thus the data size is 250,000. In case 2 where $\mu_1 \gg \mu_2$, the temporal domain changes to be $t \in [0, 1000)$ while other conditions keep the same. It is worth noting that the preasymptotic behavior is investigated here. As illustrated in Fig. 4a, the dimensionless variable is employed to discover a general PDE. We randomly select 200,000 data to train the neural network.

In this actual complex problem, some techniques are adopted to facilitate the discovery of governing equations. Firstly, considering that the system is conservative, the underlying governing equations can be transformed into the conservative form:

$$\frac{\partial h^*}{\partial t^*} + \frac{\partial F^*(x^*, t^*)}{\partial x^*} = 0, \quad (27)$$

where $F^*$ is the dimensionless flux related to $h^*$ and its derivatives. Therefore, the flux $F$ can be calculated by integration techniques as follows:

$$F^*(x^*, t^*) = \int_{x_0}^{x^*} -\frac{\partial h^*}{\partial t^*} dx^*. \tag{28}$$

Benefiting from the neural network, the $F^*$ on the meta-data can be calculated easily. Therefore, the problem is converted to discover the PDE form of the dimensionless flux $F^*$, which is accomplished by PIC in this work.

**Data availability**

All datasets utilized in this work are available on Github. The URL is https://github.com/woshixuhao/PIC_code/tree/main/data.

**Code availability**

The source code is available on Github. The URL is https://github.com/woshixuhao/PIC_code.

**Acknowledgments**

This work is partially funded by the National Center for Applied Mathematics Shenzhen (NCAMS), the Shenzhen Key Laboratory of Natural Gas Hydrates (Grant No. ZDSYS20200421111201738), and the SUSTech - Qingdao New Energy Technology Research Institute.

**References**

1. Schaeffer, H. Learning partial differential equations via data discovery and sparse optimization. *Proc. R. Soc. A Math. Phys. Eng. Sci.* **473**, (2017).
2. Rudy, S. H., Brunton, S. L., Proctor, J. L. & Kutz, J. N. Data-driven discovery of partial differential equations. *Sci. Adv.* **3**, (2017).
3. Brunton, S. L., Proctor, J. L., Kutz, J. N. & Bialek, W. Discovering governing equations from data by sparse identification of nonlinear dynamical systems. *Proc. Natl. Acad. Sci. U. S. A.* **113**, 3932–3937 (2016).
4. Fukami, K., Murata, T., Zhang, K. & Fukagata, K. Sparse identification of nonlinear dynamics with low-dimensionalized flow representations. *J. Fluid Mech.* **926**, (2021).
5. Kaheman, K., Kutz, J. N. & Brunton, S. L. SINDy-PI: A robust algorithm for parallel implicit sparse identification of nonlinear dynamics: SINDy-PI. *Proc. R. Soc. A Math. Phys. Eng. Sci.* **476**, (2020).
6. Messenger, D. A. & Bortz, D. M. Weak SINDy for partial differential equations. *J. Comput. Phys.* **443**, 110525 (2021).
7. Boninsegna, L., Nüske, F. & Clementi, C. Sparse learning of stochastic dynamical equations. *J. Chem. Phys.* **148**, (2018).
8. Brunton, S. L. & Kutz, J. N. Methods for data-driven multiscale model discovery for materials. *J. Phys. Mater.* **2**, 044002 (2019).
9. Rudy, S., Alla, A., Brunton, S. L. & Kutz, J. N. Data-driven identification of parametric partial differential equations. *SIAM J. Appl. Dyn. Syst.* **18**, 643–660 (2019).
10. Maslyaev, M., Hvatov, A. & Kalyuzhnaya, A. Data-driven partial derivative equations discovery with evolutionary approach. *Lect. Notes Comput. Sci. (including Subser. Lect. Notes*

*Artif. Intell. Lect. Notes Bioinformatics)* **11540 LNCS**, 635–641 (2019).

11. Xu, H., Chang, H. & Zhang, D. DLGA-PDE: Discovery of PDEs with incomplete candidate library via combination of deep learning and genetic algorithm. *J. Comput. Phys.* **418**, 109584 (2020).

12. Chen, Y., Luo, Y., Liu, Q., Xu, H. & Zhang, D. Symbolic genetic algorithm for discovering open-form partial differential equations (SGA-PDE). *Phys. Rev. Res.* **4**, (2022).

13. Xu, H., Chang, H. & Zhang, D. Dl-pde: Deep-learning based data-driven discovery of partial differential equations from discrete and noisy data. *Commun. Comput. Phys.* **29**, 698–728 (2021).

14. Berg, J. & Nyström, K. Data-driven discovery of PDEs in complex datasets. *J. Comput. Phys.* **384**, 239–252 (2019).

15. Zhang, Z. & Liu, Y. Robust data-driven discovery of partial differential equations under uncertainties, arxiv: 2102.06504.

16. Raissi, M. Deep hidden physics models: Deep learning of nonlinear partial differential equations. *J. Mach. Learn. Res.* **19**, 1–24 (2018).

17. Stephany, R. & Earls, C. PDE-READ: Human-readable partial differential equation discovery using deep learning, arxiv: 2111.00998.

18. Xu, H. & Zhang, D. Robust discovery of partial differential equations in complex situations. *Phys. Rev. Res.* **3**, 1–12 (2021).

19. Chen, Z., Liu, Y. & Sun, H. Physics-informed learning of governing equations from scarce data. *Nat. Commun.* **12**, 1–13 (2021).

20. Lejarza, F. & Baldea, M. DySMHO: Data-driven discovery of governing equations for dynamical systems via moving horizon optimization, arxiv: 2108.00069.

21. Raissi, M., Perdikaris, P. & Karniadakis, G. E. Physics-informed neural networks: A deep learning framework for solving forward and inverse problems involving nonlinear partial differential equations. *J. Comput. Phys.* **378**, 686–707 (2019).

22. Boullé, N., Nakatsukasa, Y. & Townsend, A. Rational neural networks, arxiv: 2004.01902.

23. Both, G. J., Choudhury, S., Sens, P. & Kusters, R. DeepMoD: Deep learning for model discovery in noisy data. *J. Comput. Phys.* **428**, 109985 (2021).

24. Burnham, K. P. & Anderson, D. R. Multimodel inference: Understanding AIC and BIC in model selection. *Sociological Methods and Research*. **33**, 261–304 (2004).

25. Xu, H., Zhang, D. & Zeng, J. Deep-learning of parametric partial differential equations from sparse and noisy data. *Phys. Fluids.* **33**, 1–30 (2021).

26. Gardner, G. H. F., Downie, J. & Kendall, H. A. Gravity segregation of miscible fluids. *SPE J.* **185**, 95-104 (1964).

27. Zeng, J., Xu, H., Chen, Y. & Zhang, D. Deep-learning discovers macroscopic governing equations for viscous gravity currents from microscopic simulation data, arxiv:2106.00009.

28. Rawlings, J. B. & Bakshi, B. R. Particle filtering and moving horizon estimation. *Comput. Chem. Eng.* **30**, (2006).

29. Allen, S. M. & Cahn, J. W. A microscopic theory for antiphase boundary motion and its application to antiphase domain coarsening. *Acta Metall.* **27**, (1979).

30. Basdevant, C. *et al.* Spectral and finite difference solutions of the Burgers equation. *Comput. Fluids* **14**, 23–41 (1986).

31. Dontsov, E. V. & Peirce, A. P. Proppant transport in hydraulic fracturing: Crack tip screen-out

in KGD and P3D models. *Int. J. Solids Struct.* **63**, (2015).

**Extended Data**

**Extended Data Table 1. The discovered PDE by PIC for the parametric convection-diffusion equation.**

| Noise level | Discovered equation |
|---|---|
| 0% | $u_t = -(0.999 + 0.252\sin(1.00x))u_x + 0.995u_{xx}$ |
| 25% | $u_t = -(0.983 + 0.228\sin(0.99x))u_x + 0.996u_{xx}$ |
| 50% | $u_t = -(1.012 + 0.298\sin(1.059x))u_x + 0.992u_{xx}$ |
| 75% | $u_t = -(1.17e^{-0.034x} + 0.269\cos(-1.37x))u_x + 0.919xe^{-0.312x}u_{xx}$ |

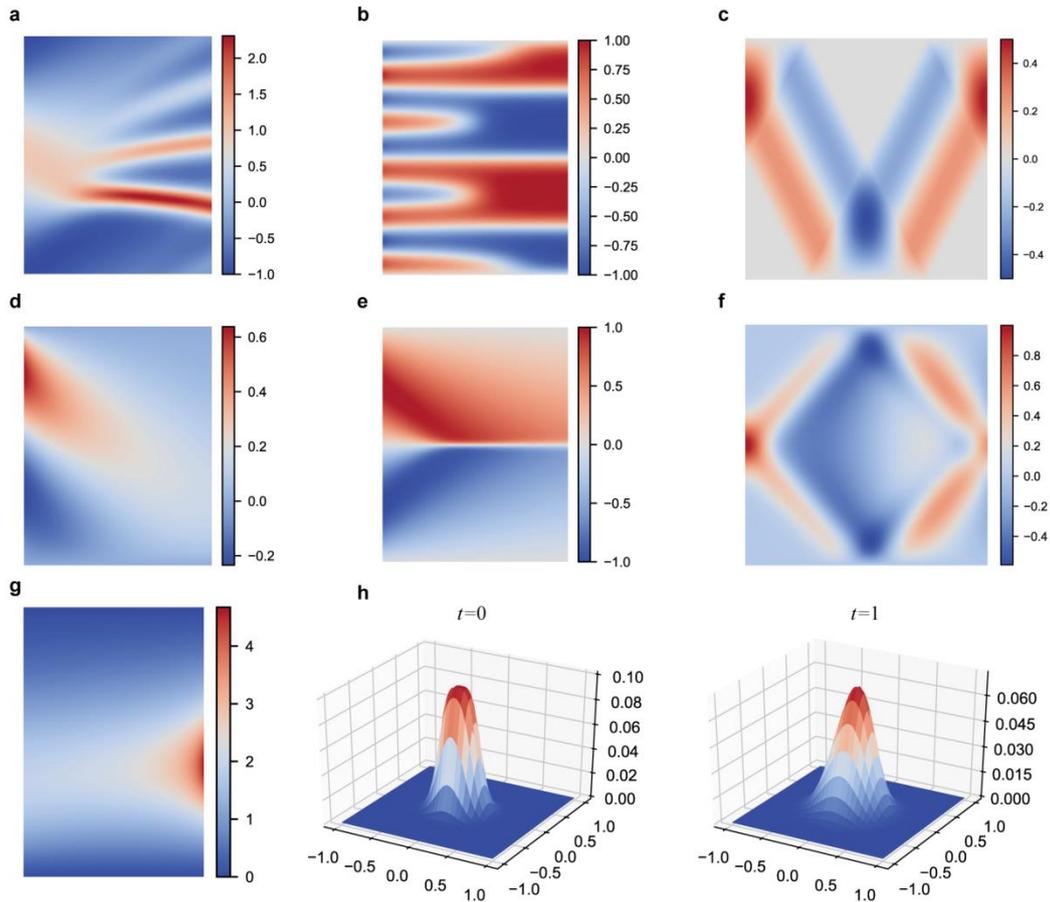

**Extended Data Fig. 1. Visualization of the dataset for seven canonical PDEs (a~g) and the 2D Burgers equation (h)**. **a**, The Korteweg-De Vries (KdV) equation; **b**, The Allen-Chan equation; **c**, The wave equation; **d**, The convection-diffusion equation; **e**, The Burgers equation; **f**, The Klein-Gordon (KG) equation; **g**, The Chaffee-infante equation; **h**, The 2D Burgers equation. For 1D cases (**a** to **g**), the heatmap of $u(x,t)$ is provided, where $x$ is on the left and $t$ is on the bottom. For the

subfigure **h**, the 2D surface of $u(x,y)$ when $t=0$ and $t=1$ is provided, where $x$ is on the left and $y$ is on the right.

# Supplementary Information for

# Discovery of partial differential equations from highly noisy and sparse data with physics-informed information criterion

Hao Xu[a], Junsheng Zeng[b], and Dongxiao Zhang[c,d,*]

[a] *BIC-ESAT, ERE, and SKLTCS, College of Engineering, Peking University, Beijing 100871, P. R. China*

[b] *Institute of Applied Physics and Computational Mathematics, Beijing 100088, P. R. China*

[c] *National Center for Applied Mathematics Shenzhen (NCAMS), Southern University of Science and Technology, Shenzhen 518055, Guangdong, P. R. China*

[d] *Department of Mathematics and Theories, Peng Cheng Laboratory, Shenzhen 518000, Guangdong, P. R. China*

[*] Corresponding author.

F-mail address: 390260267@pku.edu.cn (H. Xu); zengjs1993@163.com (J. Zeng); zhangdx@sustech.edu.cn (D. Zhang).

## 1. Experimental settings

For the reproducibility of this work, we provide detailed experimental settings, including information about the methods, parameters, and data in this section.

### 1.1 The settings of the neural network and physics-informed neural network

This work uses a fully connected artificial neural network (ANN) to construct a surrogate model, generate meta-data, and calculate derivatives via automatic differentiation. For the ANN, the number of hidden layers is 5, with 50 neurons in each hidden layer, the number of input neurons is 2, and the number of output neurons is 1. The input is the spatial-temporal location ($x,t$), and the target is the observation $u$. The activation function is the Sin function[1,2], $f(x)=\sin(x)$, or the Rational function[3], which is written as:

$$f(x) = \frac{P(x)}{Q(x)} = \frac{\sum_{i=1}^{r_P} a_i x^i}{\sum_{j=1}^{r_Q} b_j x^j}, \qquad (1)$$

where $P(x)$ and $Q(x)$ are polynomials of $x$ with the order of $r_P$ and $r_Q$, respectively. In this work, $r_P$ and $r_Q$ are 3 and 2, respectively. The Rational activation function is proven to be more suitable for some cases with high noise[4]. Therefore, in this work, we compare the result of both activation functions and select the best one for each case. The optimizer is chosen to be Adam with a learning rate of 0.001. The maximum training epoch is 30,000, and the early stop technique is adopted to prevent overfitting based on the training loss and validating loss. The construction of the physics-informed neural network (PINN) is the same as the ANN, with different loss functions mentioned in the Methods section of the main text. For calculating the physical loss (*p*-loss), the PINN is trained for 300 epochs based on the pre-trained neural network for each combination. The $\lambda_{Data}$ and $\lambda_{PDE}$ are 1 and 0.01, respectively, which are decided empirically. After the ultimate PDE is determined, the coefficients are optimized by PINN. Here, the training epoch of PINN is 3,000 for the KG equation, the wave equation, and 1,000 for the others.

### 1.2 The settings of the meta-data

Benefitting from the neural network, we can construct a surrogate model from a few sparse, noisy data to generate meta-data on grids, which plays a vital role in the generalized genetic algorithm and calculation of *r*-loss and *p*-loss. Here, we provide information on the meta-data for the PDEs utilized in this work. In all cases, the meta-data for the generalized genetic algorithm and *r*-loss are grid data of 100 spatial observation points and 100 temporal observation points; thus, the data size is 10,000. For the PINN, the test data for calculating *p*-loss are grid data of 100×100 for the Rational activation function and 200×200 for the Sin activation function. The domain of the meta-data in each PDE is detailed in Table S1.

**Table S1. The spatial and temporal domains of the meta-data for the canonical PDEs utilized in this work.**

| Equation name | Equation form | Spatial domain | Temporal domain |
|---|---|---|---|
| KdV equation | $u_t = -uu_x - 0.0025u_{xxx}$ | $x' \in [-0.8, 0.8]$ | $t' \in [0.1, 0.9]$ |
| Burgers equation | $u_t = -uu_x + 0.1u_{xx}$ | $x' \in [-7, 7]$ | $t' \in [1, 9]$ |
| Convection-diffusion equation | $u_t = -u_x + 0.25u_{xx}$ | $x' \in [0, 2]$ | $t' \in [0, 1]$ |
| Chaffee-Infante equation | $u_t = u_{xx} - u + u^3$ | $x' \in [0.5, 2.5]$ | $t' \in [0.15, 0.45]$ |
| Allen-Cahn equation | $u_t = 0.003u_{xx} + u - u^3$ | $x' \in [-0.8, 0.8]$ | $t' \in [1, 9]$ |
| Wave equation | $u_{tt} = u_{xx}$ | $x' \in [0.1, 3]$ | $t' \in [0.2, 6]$ |
| KG equation | $u_{tt} = 0.5u_{xx} - 5u$ | $x' \in [-0.8, 0.8]$ | $t' \in [0.3, 2.7]$ |

It can be seen that the domain of meta-data is slightly narrower than that of the observation data, which is because the derivatives near the boundary are relatively inaccurate. Therefore, we avoided the original boundary when generating meta-data.

### 1.3 The settings of the moving horizon

When obtaining the *r*-loss of the combinations, the smoothed meta-data are divided into $N_h$ overlapping horizons $T_i$, which is defined as $[t'_{min} + i\Delta t, \frac{1}{2}(t'_{min} + t'_{max}) + i\Delta t]$, where $t'_{min}$ and $t'_{max}$ are the minimum and maximum of the time domain of the meta-data $t'$, respectively; $i=1,2,...,N_h$; and $\Delta t$ is the length of horizons. In this work, $\Delta t$ is set to be:

$$\Delta t = \frac{t'_{max} - t'_{min}}{2 \cdot N_h}. \tag{2}$$

In this work, $N_h$ is 10.

**1.4 The settings for the generalized genetic algorithm**

In this section, we provide the settings for the generalized genetic algorithm. In this work, the population size of genomes is 400, and the number of maximum generations is 200. The rate of cross-over is 1.0, the rate of add mutation and the basic gene mutation is 0.4, and the rate of delete module mutation is 0.5. The basic genes are $u$, $u_x$, $u_{xx}$ and $u_{xxx}$, and the derivative order is considered up to 3. This means that the maximum derivative order for terms can be up to 6 (e.g., $\frac{\partial^3 u_{xxx}}{\partial x^3}$), which is sufficient for most situations. The $l_0$ penalty in the fitness is 0.1 for the KdV equation, the Chaffee-Infante equation and the KG equation, 0.01 for the wave equation and the convection-diffusion equation, and 0.0005 for the Burgers equation and the Allen-Cahn equation. The hyper-parameter $l_0$ penalty is selected according to trial and error, and the selection is not restricted (proved in the main text). A proper $l_0$ penalty leads to a preliminary library with a suitable number of terms (usually lower than 10 terms), which will conserve the calculation cost when calculating $r$-loss.

**1.5 The setting of the experiment examining the extendibility of the PIC**

In this work, the 2D Burgers equation and the parametric convection-diffusion equation are employed to show the extendibility of the proposed PIC. Here, the information about these two PDEs is provided.

For the 2D Burgers equation, the dataset is the same as that utilized in Zhang and Liu[5] where the initial condition is $u(x,y,0) = 0.1\text{sech}(20x^2 + 25y^2)$, and the boundary condition is periodic. The dataset is the grid data of 101 spatial observation points in the domain $x \in [-1,1]$, 51 spatial observation points in the domain $y \in [-1,1]$, and 100 temporal observation points in the domain $t \in [0,2)$. The data size is 515,100. The meta-data are the grid data of 20 spatial observation points in the domain $x \in [-0.8,0.8]$, 20 spatial observation points in the domain $y \in [-0.8,0.8]$, and 20 temporal observation points in the domain $t \in [0,2]$. The meta-data size is 8,000. In this work, 200,000 discrete data are randomly selected from the dataset to construct the training data.

For the parametric convection-diffusion equation, the form is written as:

$$u_t = -(1+0.25\sin(x))u_x + u_{xx}. \qquad (3)$$

The dataset is the same as that in Xu et al.[6], which has 250 temporal observation steps in $t \in [0,5)$ and 201 spatial observation steps in $x \in [0,8]$; thus, the data size is 50,250. The meta-data are grid data of 100 spatial observation points in the domain $x \in [0.5, 7.5]$, and 100 temporal observation points in the domain $t \in [0.2, 4.8]$. In this work, 30,000 discrete data are randomly selected from the dataset to construct the training data. The discovery of parametric PDEs by PIC is based on previous work[6], which utilizes stepwise methods to identify the PDE structure and then employs the PINN to calculate the variant coefficients. Finally, the expression of coefficients is discovered by the DLGA-PDE (coefficients)[6]. In this work, a preliminary library can be obtained from the generalized genetic algorithm in each local window, which can be combined into a summarized library for the global process. Then, the *r*-loss of all possible combinations of the summarized library is calculated, and the *p*-loss of several top combinations is calculated. The PDE loss is calculated by:

$$L_{PINN}(\theta) = \lambda_{Data} MSE_{Data} + \lambda_{PDE} MSE_{PDE},$$

$$MSE_{Data} = \frac{1}{N_x N_t} \sum_{i=1}^{N_x} \sum_{j=1}^{N_t} (u(x_i, t_j) - PINN(x_i, t_j; \theta)), \qquad (4)$$

$$MSE_{PDE} = \frac{1}{N'_x N'_t} \sum_{i=1}^{N'_x} \sum_{j=1}^{N'_t} (U'_L(x'_i, t'_j) - \xi'_i \cdot U'_R(x'_i, t'_j))^2.$$

The difference is that the $\xi'_i$ in the $MSE_{PDE}$ is a variable, and the value on each *x* is calculated by solving the optimal solution of $U'_L(x'_i) - \xi'_i \cdot U'_R(x'_i) = 0$ by least squared regression.

## 2. Supplementary Experiments

### 2.1 The influence of data noise

In this work, the robustness of the proposed PIC to data noise is examined, and satisfactory results are obtained in which the PIC is robust to high levels of Gaussian noise with sparse, discrete data. In this section, we investigate the performance of PIC with different types of data, including Gaussian noise, uniform noise, and random field noise. Gaussian noise is a commonly used noise type in previous works[5], while uniform noise is another noise type in PDE discovery works[7]. Both Gaussian and uniform noise are white noise. In contrast, random field noise is a non-white noise rarely seen in previous works, which is more challenging for PDE discovery and is more realistic. The KdV equation is taken as an example here, and the maximum noise to which PIC can be robust is illustrated in Fig. S1.

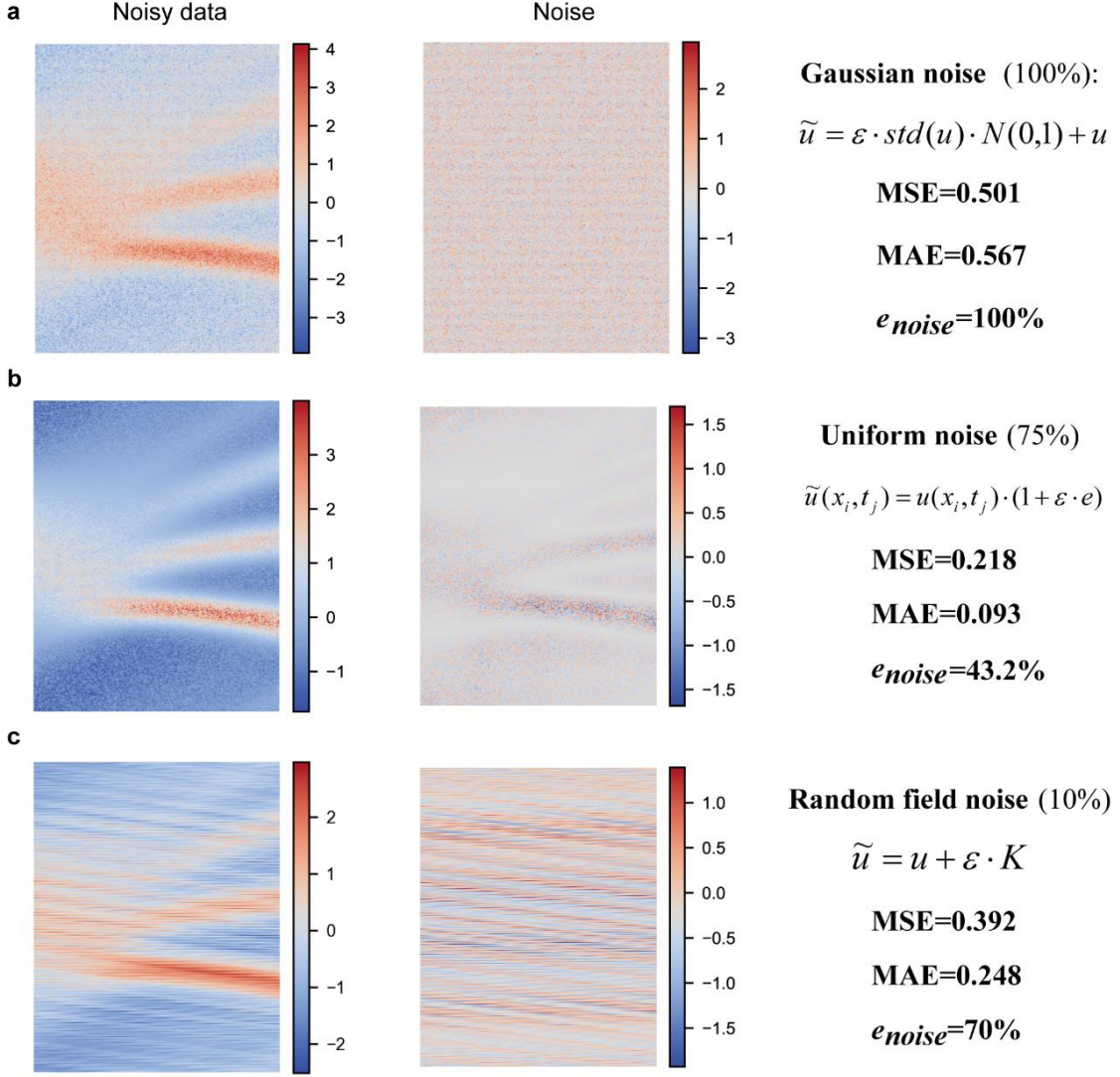

**Fig. S1. Comparison between different types of noise. a,** Gaussian noise. **b,** Uniform noise. **c,** Random field noise. The KdV equation is taken as an example here, and the noise level is the maximum noise that PIC can handle. The left is the noisy data, and the middle is the visualization of noise. $K$ is a Gaussian random field with a correlation length of 32.

In order to better compare these types of noise, the MSE, MAE, and relative error of the noise $e_{noise}$ are calculated as follows:

$$MSE = \frac{1}{N_t N_x} \sum_{i=1}^{N_x} \sum_{j=1}^{N_t} (u(x_i, t_j) - \tilde{u}(x_i, t_j))^2,$$

$$MAE = \frac{1}{N_t N_x} \sum_{i=1}^{N_x} \sum_{j=1}^{N_t} |u(x_i, t_j) - \tilde{u}(x_i, t_j)|, \quad (5)$$

$$e_{noise} = \left( \frac{\sum_{i=1}^{N_x}\sum_{j=1}^{N_t}(u(x_i,t_j)-\tilde{u}(x_i,t_j))^2}{\sum_{i=1}^{N_x}\sum_{j=1}^{N_t}(u(x_i,t_j))^2} \right)^{\frac{1}{2}} \times 100\%.$$

where $u$ are the clean data; $\tilde{u}$ are the noisy data; and $N_x$ and $N_t$ are the number of $x$ and $t$ of the data, respectively. From the figure, it is found that the PIC can handle these three types of noise well, which confirms the robustness of PIC to different types of noise. Meanwhile, it can be seen that the MSE, MAE and $e_{noise}$ of the Gaussian noise are the largest among these three types of noise, which means that the PIC is more suitable to handle Gaussian noise. It is interesting to find that the uniform noise is harder to deal with than the Gaussian noise. The reason for this may be that the uniform noise greatly influences the major part of the physical process. In contrast, the Gaussian process affects the entire space-time domain, which is easier to smooth by the neural network.

**2.2 The effectiveness of the generalized genetic algorithm**

The generalized genetic algorithm is utilized in this work to obtain a preliminary potential library. It is crucial for the preliminary library to contain all correct terms to be a complete library. Therefore, the ability of the generalized genetic algorithm to guarantee the correct terms to be included should be examined. In previous research[8], it has been proven that high-order redundant compensation terms can be discovered by the generalized genetic algorithm, which can compensate for the error from noise and guarantee the inclusion of the correct dominant terms. Here, we also conduct an experiment to confirm this issue. Here, the KdV equation with 100% Gaussian noise is taken as an example, and the results are provided in Table S2. The table shows that the coefficients calculated by direct least square regression on the true terms derived a lot from the true coefficients. In contrast, although the generalized genetic algorithm discovered a redundant term $u_x u_{xx}$, the coefficients of the true terms ($uu_x$ and $u_{xxx}$) are accurate, which shows that the existence of the redundant term can compensate the error brought by the high noise and make the true terms accurate. This is why the generalized genetic algorithm can guarantee the correct terms to be included under high noise levels.

**Table S2. Comparison between the true PDE and the outcomes from the generalized genetic algorithm, direct least square regression on the true terms, and PIC.**

| | |
|---|---|
| The true PDE | $u_t = -uu_x - 0.0025u_{xxx},$ |
| Direct least square regression on the true terms | $u_t = -0.490uu_x - 0.0011u_{xxx}$ |
| The discovered PDE by the generalized genetic algorithm | $u_t = -0.930uu_x - 0.0027u_{xxx} - 0.0015u_x u_{xx}$ |
| Ultimate discovered PDE by PIC | $u_t = -1.043uu_x - 0.0026u_{xxx}$ |

**2.3 The variation of coefficients in the moving horizon**

In this work, the moving horizon technique is utilized to calculate *r*-loss and measure the parsimony of the discovered PDE. In order to better demonstrate the function of the moving horizon technique, we visualize the variation of coefficients in the moving horizon in Fig. S2. Here, three types of combinations are illustrated, including the true PDE (Fig. S2a), the false PDE (Fig. S2b), and the true PDE with redundant terms (Fig. S2c). It is evident that the mean *cv* of the true PDE is the smallest, while the mean *cv* is much larger for the true PDE with redundant terms. The main reason for this is the high variance (*cv*=0.144) from the redundant term $uu_{xxx}$. This proves the validity of the moving horizon technique to distinguish the redundant terms, which allows the *r*-loss to measure the parsimony of the discovered PDE.

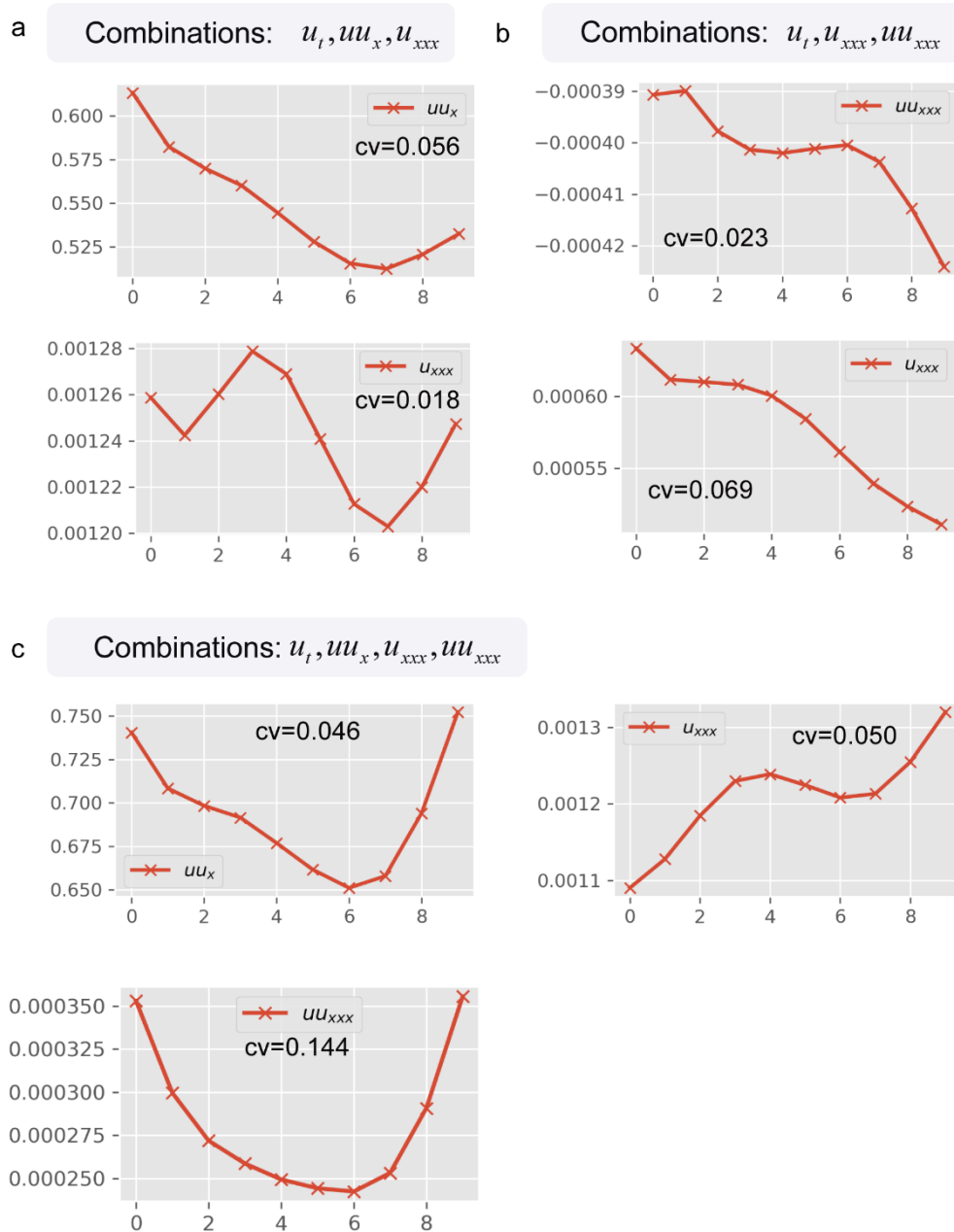

**Fig. S2. The variation of coefficients in the moving horizon for different combinations. a,** The true PDE. **b,** The false PDE. **c,** The true PDE with redundant terms. The KdV equation under 100%

Gaussian noise is taken as an example here. The *x*-axis represents different horizons, and the *y*-axis represents corresponding coefficients.

**References**
1. Raissi, M. Deep hidden physics models: Deep learning of nonlinear partial differential equations. *J. Mach. Learn. Res.* **19**, 1–24 (2018).
2. Raissi, M., Perdikaris, P. & Karniadakis, G. E. Physics-informed neural networks: A deep learning framework for solving forward and inverse problems involving nonlinear partial differential equations. *J. Comput. Phys.* **378**, 686–707 (2019).
3. Boullé, N., Nakatsukasa, Y. & Townsend, A. Rational neural networks, arxiv: 2004.01902.
4. Stephany, R. & Earls, C. PDE-READ: Human-readable partial differential equation discovery using deep learning, arxiv: 2111.00998.
5. Zhang, Z. & Liu, Y. Robust data-driven discovery of partial differential equations under uncertainties, arxiv: 2004.01902.
6. Xu, H., Zhang, D. & Zeng, J. Deep-learning of parametric partial differential equations from sparse and noisy data. *Phys. Fluids* **33**, 0–17 (2021).
7. Chang, H. & Zhang, D. Machine learning subsurface flow equations from data. *Comput. Geosci.* **23**, 895–910 (2019).
8. Xu, H. & Zhang, D. Robust discovery of partial differential equations in complex situations. *Phys. Rev. Res.* **3**, 1–12 (2021).